\documentclass[11pt,a4paper]{article}

\usepackage[utf8]{inputenc}
\usepackage[T1]{fontenc}
\usepackage{lmodern}
\usepackage{geometry}
\usepackage{amssymb}
\usepackage{hyperref}
\usepackage{listings}
\usepackage{xcolor}
\usepackage{graphicx}
\usepackage{subcaption}
\usepackage[numbers,sort&compress]{natbib}

\lstdefinestyle{code}{
    backgroundcolor=\color{gray!10},
    basicstyle=\ttfamily\footnotesize,
    frame=single,
    breaklines=true,
    numbers=left,
    numberstyle=\tiny\color{gray},
    keywordstyle=\color{blue},
    stringstyle=\color{teal},
    commentstyle=\color{gray}\itshape,
    tabsize=2,
    captionpos=b
}

\title{Robodimm: A Physics-Grounded Framework for Automated Actuator Sizing in Scalable Modular Robots}

\author{
J.L. Torres\textsuperscript{1},
M. Mu\~noz\textsuperscript{2},
J.D. \'Alvarez\textsuperscript{2},
J.L. Blanco\textsuperscript{1},
A. Gimenez\textsuperscript{1}\\
\small \textsuperscript{1}CIESOL-ceiA3, Department of Engineering, University of Almer\'ia, Spain\\
\small \textsuperscript{2}CIESOL-ceiA3, Department of Computer Science, University of Almer\'ia, Spain\\
\small Corresponding author: \href{mailto:jltmoreno@ual.es}{jltmoreno@ual.es}
}

\date{}

\begin{document}

\maketitle

\begin{abstract}
	Selecting an appropriate motor--gearbox combination is a critical design task in robotics because it directly affects cost, mass, and dynamic performance. This process is especially challenging in modular robots with closed kinematic chains, where joint torques are coupled and actuator inertia propagates through the mechanism. We present Robodimm, a software framework for automated actuator sizing in scalable robot architectures. By leveraging Pinocchio for dynamics and Pink for inverse kinematics, Robodimm uses a Karush--Kuhn--Tucker (KKT) formulation for constrained inverse dynamics. The platform supports parametric scaling, interactive trajectory programming through jog modes, and a two-round validation workflow that addresses actuator self-weight effects.
\end{abstract}

\textbf{Keywords:} Actuator sizing; inverse dynamics; closed kinematic chains; palletizing robots; SMEs; web-based robotics software.

\section{Introduction}

Robotic systems are now central to manufacturing, logistics, agriculture, inspection, and service applications. In all these domains, simulation tools are essential for safe and cost-effective development, testing, and validation. Realistic simulation environments allow engineers to explore alternative system configurations, evaluate control strategies, and reduce the time and cost of physical prototyping. In mobile robotics, simulators such as MVSim~\cite{BLANCOCLARACO2023101443} have demonstrated the value of accessible and physically grounded frameworks for both research and education. In industrial manipulation, software frameworks such as~\cite{elsner2023panda} have simplified robot programming by providing high-level interfaces that support integration with modern perception and learning pipelines. Likewise, platforms such as~\cite{barron2026mobile} illustrate the role of simulation in broadening access to robotics in STEM/STEAM contexts.

Beyond mobile platforms, industrial manipulators are a cornerstone of modern automation. They are widely used in tasks requiring precision, repeatability, and reliability, including assembly, material handling, welding, and inspection. In these settings, simulation is no longer limited to geometric checks or collision avoidance: dynamic behavior under realistic operating conditions must also be assessed. Payload variation, motion profile shape, actuator limits, and inter-joint coupling all influence performance, safety, and energy efficiency.

Despite the wide availability of simulation tools, many manipulator workflows still focus mainly on kinematic modeling and motion planning. Dynamic analysis and actuator sizing are often possible only by combining external libraries, custom scripts, and specialized tooling. This fragmentation increases engineering overhead, raises the barrier for non-expert users, and reduces the practical use of dynamics in early design stages. The problem is more pronounced in closed-chain manipulators, such as parallelogram mechanisms, where coupled joints make open-chain approximations physically inconsistent.

General-purpose ecosystems provide strong foundations for robot software and virtual experimentation. Examples include ROS~2~\cite{macenski2022ros2}, Gazebo~\cite{koenig2004gazebo}, and V-REP/CoppeliaSim~\cite{rohmer2013vrep}. However, actuator sizing for closed-chain manipulators still typically depends on custom dynamic pipelines and manual iteration outside a unified user-facing workflow. On the dynamics side, rigid-body libraries such as Pinocchio~\cite{pinocchio2019,carpentier2018pinocchio} and IK tools such as Pink~\cite{pink_github} make high-performance constrained computation feasible, but translating those capabilities into an accessible end-to-end sizing process remains a practical gap.

This work addresses these challenges by introducing Robodimm, a software platform specifically designed to support the simulation and dynamic analysis of industrial robotic manipulators with closed kinematic chains. By integrating interactive motion definition, constrained inverse dynamics, and actuator-related analysis into a unified web environment, Robodimm abstracts the underlying computational complexity and provides a transparent user experience.

Compared with general-purpose simulators, Robodimm is intentionally scoped to trajectory-to-sizing studies for manipulator design loops. It emphasizes task-level programming, constrained torque estimation under mechanism closure, and catalog-based actuator recommendation in one workflow rather than full-scene physics or broad plugin ecosystems.

The practical motivation is especially strong for palletizing robots, where many industrial tasks can be solved with 4 actuated joints and a closed-chain architecture instead of a full 6-DOF arm. This reduces control complexity, hardware cost, and commissioning effort while preserving useful workspace and throughput for pick-and-place workflows.

\section{System Overview}

Robodimm is a web-based platform for modeling, simulation, and dynamic analysis of industrial manipulators, with particular focus on closed-chain mechanisms such as parallelogram structures. It integrates interactive visualization, motion generation, and constrained dynamic computation in a single browser-accessible environment.

\subsection{Software architecture}

\begin{figure}[h!]
	\centering
	\includegraphics[width=0.95\linewidth]{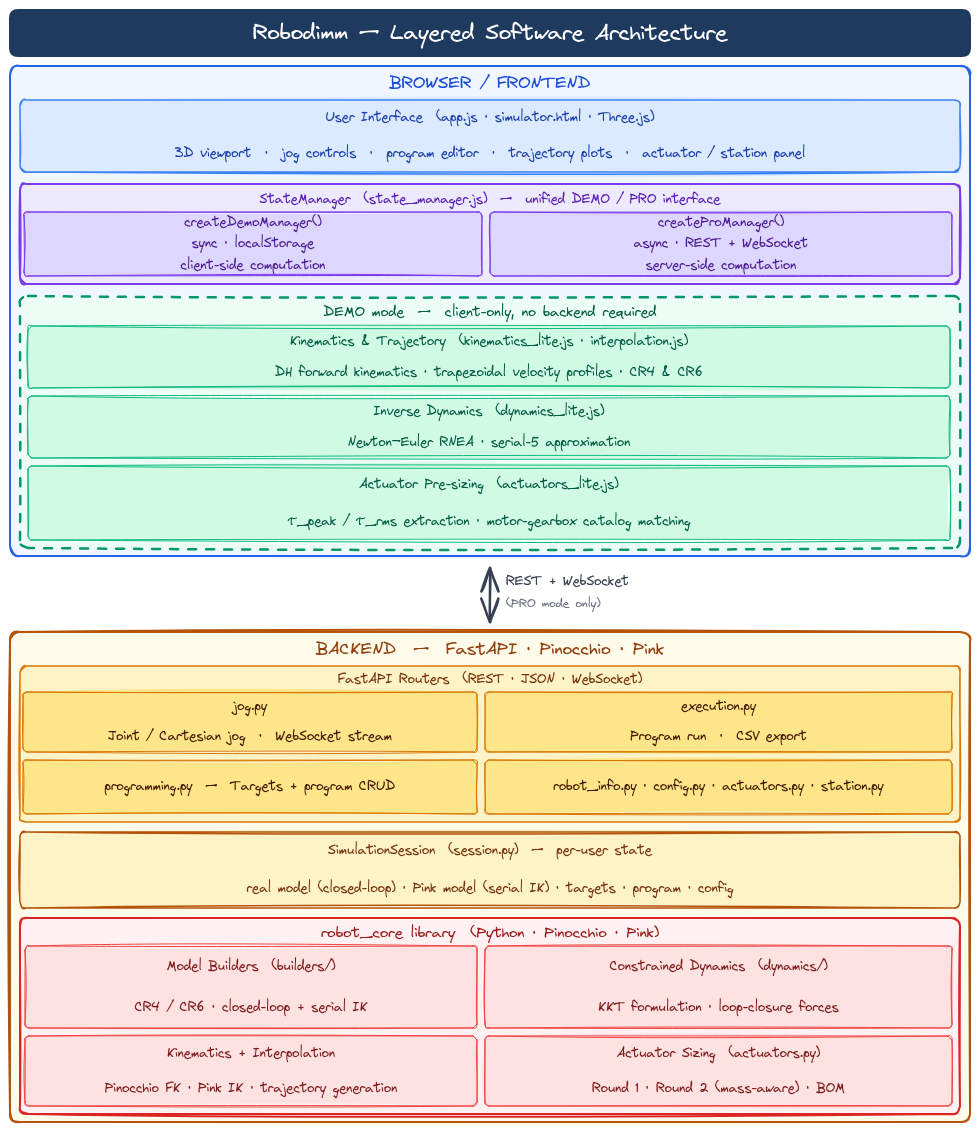}
	\caption{Unified Robodimm architecture with DEMO and PRO execution paths.}
	\label{fig:Architecture}
\end{figure}

Robodimm is designed following a layered architecture that separates presentation, application logic, and computational responsibilities. Figure~\ref{fig:Architecture} summarizes the unified DEMO/PRO design, including shared state management, backend services, and the computational core.

\begin{itemize}
	\item \textbf{Frontend:} Runs in a standard web browser and provides interfaces for robot configuration, manual control, trajectory programming, and result inspection.
	\item \textbf{Backend:} Implemented with FastAPI. It exposes REST endpoints and a WebSocket stream for real-time robot state updates.
	\item \textbf{robot\_core library:} Encapsulates kinematics, interpolation, constrained inverse dynamics, and actuator analysis.
\end{itemize}

\subsection{Software functionalities}

Robodimm provides an integrated workflow from robot configuration and motion definition to trajectory-level dynamic analysis and actuator sizing.

\begin{itemize}
	\item \textbf{Dual execution modes:} A \emph{DEMO} mode (frontend-only) enables rapid tests without backend dependencies, while a \emph{PRO} mode (backend-driven) provides constrained inverse dynamics, station integration, and actuator sizing.
	\item \textbf{Robot configuration and modeling:} Parametric CR4 (4-DOF palletizer) and CR6 (6-DOF manipulator) models, both supported by the Robodimm workflow in DEMO and PRO modes.
	\item \textbf{Motion definition and kinematics:} Joint and Cartesian jog, waypoint programming, and forward/inverse kinematics.
	\item \textbf{Closed-chain inverse dynamics:} KKT-based constrained dynamics for physically consistent torque estimation in coupled mechanisms.
	\item \textbf{Actuator requirement extraction:} Peak and RMS torque/speed extraction for motor-gearbox selection.
	\item \textbf{Visualization and station integration:} Real-time 3D view and station geometry import.
\end{itemize}

\subsection{Graphical user interface}

Robodimm is operated through a single-page web interface organized into functional panels: robot/station setup, jog controls, waypoint and program editing, trajectory execution, and post-run plots for torque and kinematic signals. The same interface supports both DEMO and PRO modes, preserving the interaction flow while changing only the computation backend.

\begin{figure}[ht]
	\centering
	\includegraphics[width=0.95\linewidth]{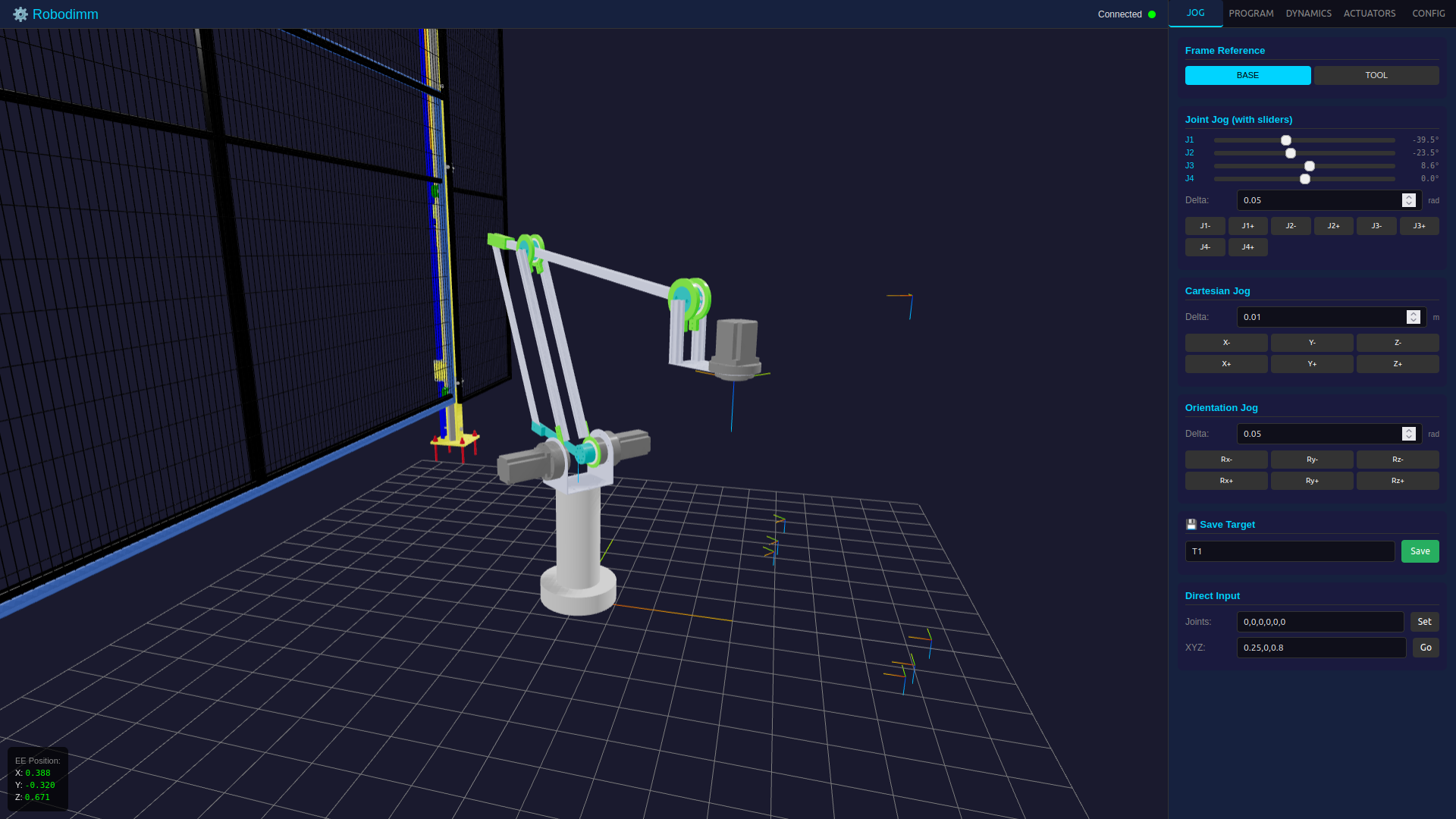}
	\caption{Robodimm graphical interface. The layout integrates 3D visualization, jog/program tools, robot configuration, and dynamic analysis outputs in a single workspace.}
	\label{fig:gui}
\end{figure}

\section{Case Study and Experimental Evaluation}

This section presents a representative CR4 palletizing example focused on endpoint motion, trajectory-level dynamics, and DEMO/PRO consistency analysis. Although Robodimm also supports CR6 6-DOF robots, the experimental case study in this article is centered on CR4 because palletizing is the target industrial application discussed here.

\subsection{Palletizing endpoint example}

The task consists of a cyclic pick-and-place motion between two pallet positions with fixed tool orientation at the endpoint. The trajectory is defined from waypoints and executed as a sequence of motion primitives. This setup is representative of high-cycle handling and stresses the coupled shoulder--elbow dynamics that dominate actuator sizing in the CR4 architecture.

The target application is end-of-line palletizing in agri-food logistics (e.g., placing tomato boxes of about 7~kg into full pallets), with a design payload margin of 10~kg for robust operation.

For reproducibility, we use the same trajectory profile and dynamic inputs in both modes: scale factor, payload mass/inertia, viscous friction, Coulomb friction, and reflected rotor inertia. In the benchmark reported here, the scenario parameters are: robot scale $s=1.6$, payload mass $10.0$ kg, payload COM $(0,0,0.1)$ m, and payload inertia diagonal $(0.0547,\,0.0963,\,0.1083)$ kg\,m$^2$ (rounded from the exact values in the scenario CSV). With the CR4 reference reach of 0.945~m at $s=1.0$, this corresponds to an approximate reach of 1.51~m. For this industrial case study, we override the default geometric-similarity exponents ($m\sim s^{3.0}$ and $I\sim s^{5.0}$) with calibrated values $m\sim s^{1.7}$ and $I\sim s^{3.7}$.

\subsection{Trajectory-based dynamic evaluation}

The example trajectory corresponds to a palletizing cycle in a CR4 robot, including pick and place waypoints with realistic velocity and acceleration profiles. The generated trajectory is evaluated in both available modes: DEMO (frontend DH/Newton--Euler pipeline) and PRO (backend Pinocchio constrained dynamics pipeline). In the shared results folder, the trajectory-generation manifest stores nominal generation settings, while the final dynamic payload/friction/inertia overrides used for torque computation are recorded in the simulation-parameter CSV.

\subsection{Actuator requirement extraction and selection}

The tool computes torque profiles along the trajectory, extracts design-relevant indicators, and supports actuator selection under payload and scaling constraints. We use a two-round process: round 1 proposes motor--gearbox pairs from trajectory requirements, while round 2 applies a mass-aware validation step to detect configuration changes or feasibility loss once actuator self-weight is considered. In the current implementation, this validation combines a fast mass-aware estimation step with optional re-simulation in PRO mode after applying the selected actuator configuration.

For the final benchmark case ($s=1.6$, payload 10~kg), the extracted peak torque requirements are 33.294~Nm (J1), 419.294~Nm (J2), 255.756~Nm (J3), and 0.426~Nm (J4), with corresponding peak joint speeds of 1.200~rad/s (11.460~rpm) for J1/J2, 1.066~rad/s (10.180~rpm) for J3, and 0~rad/s in this specific cycle for J4 (tool orientation fixed). The resulting motor--gearbox sizing remains within the current catalog for all four actuators after round-2 mass-aware validation.

\subsection{DEMO versus PRO torque comparison}

A key validation step is the comparison of both execution modes on the same programmed cycle. DEMO uses a serial-5 DH dynamic model mapped to CR4 actuated joints, while PRO uses constrained dynamics in Pinocchio. For practical design loops, we compare motor torque traces in terms of trend agreement and magnitude deviation across the full cycle.

The comparison is used as a diagnostic layer: when both traces agree sufficiently, DEMO can be used for rapid iteration; when differences increase (typically in coupled joints), PRO remains the reference for final actuator verification.

For industrial deployment, this separation is valuable: teams can iterate quickly in DEMO during concept exploration, then consolidate final sizing decisions in PRO.

\begin{table}[ht]
	\centering
	\caption{Metrics reported for DEMO/PRO torque comparison in the palletizing cycle (scale 1.6, payload 10 kg).}
	\label{tab:demo_pro_metrics}
	\begin{tabular}{lccc}
		\hline
		Joint & Correlation & RMSE [Nm] & Bias [Nm] \\
		\hline
		J1    & 1.0000      & 0.026     & -0.000    \\
		J2    & 0.9994      & 2.773     & -0.920    \\
		J3    & 0.9996      & 1.545     & -0.036    \\
		J4    & 0.9961      & 0.089     & -0.000    \\
		\hline
	\end{tabular}
\end{table}

\begin{table}[ht]
	\centering
	\caption{Two-round actuator selection results (CR4 benchmark, safety factors: torque 1.5, speed 1.2).}
	\label{tab:actuator_rounds}
	\begin{tabular}{lcc}
		\hline
		Joint & Round 1 recommendation        & Round 2 validation            \\
		\hline
		J1    & AC\_400W\_2500 + ZXS20\_50    & AC\_600W\_2500 + ZXS20\_100   \\
		J2    & AC\_2\_3kW\_1500 + ZXS40\_100 & AC\_2\_3kW\_1500 + ZXS40\_100 \\
		J3    & AC\_1000W\_2500 + ZXS40\_160  & AC\_1000W\_2500 + ZXS40\_160  \\
		J4    & AC\_200W\_3000 + ZXS14\_30    & AC\_200W\_3000 + ZXS14\_30    \\
		\hline
	\end{tabular}
\end{table}

The round-2 update in J1 (from ZXS20\_50 to ZXS20\_100) reflects the mass-aware validation step: once actuator self-weight is propagated through the mechanism, the higher reduction ratio provides additional torque margin without changing J2--J4 selections.

\begin{figure}[ht]
	\centering
	\includegraphics[width=\linewidth]{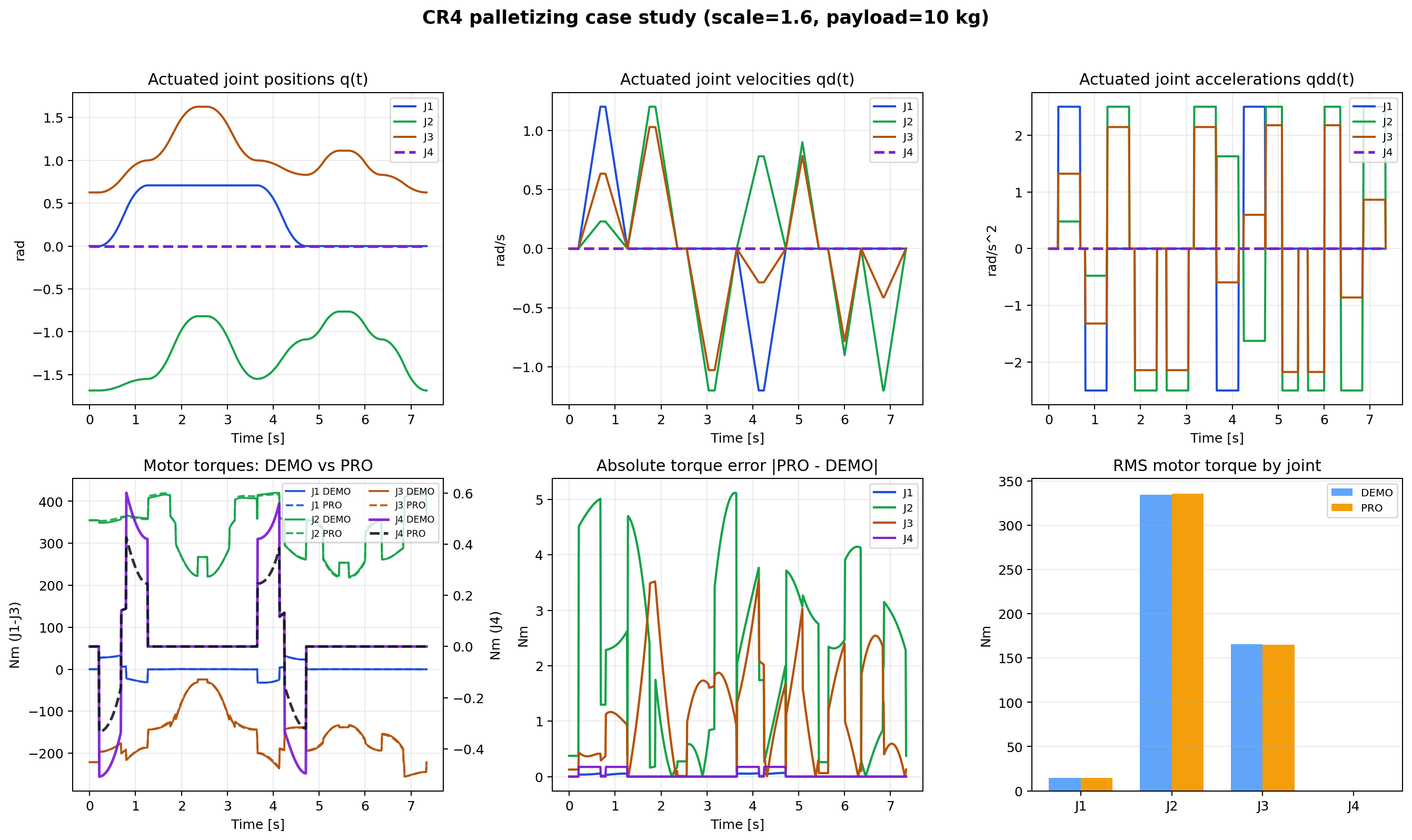}
	\caption{CR4 palletizing case study for the final benchmark scenario (scale 1.6, payload 10 kg). Top row: actuated joint positions, velocities, and accelerations. Bottom row: DEMO/PRO torque comparison, absolute torque error, and RMS torque by joint. In the torque panel, J1--J3 use the left axis and J4 uses the right axis.}
	\label{fig:trajectory_analysis}
\end{figure}

\section{Discussion and Impact}

Robodimm enables systematic studies of industrial manipulators with closed kinematic chains, where loop-closure and coupling effects can dominate actuator demand. By unifying trajectory programming, constrained inverse dynamics, and actuator sizing in a single workflow, the platform reduces manual integration effort and improves reproducibility for both research and engineering practice. In practical engineering loops, this shortens the iteration from trajectory definition to validated motor--gearbox recommendations: teams can screen many alternatives in DEMO mode and reserve PRO constrained dynamics for final verification, avoiding repeated tool handoffs across CAD exports, ad-hoc scripts, and offline spreadsheets.

From a technology-transfer perspective, the framework is particularly relevant for SMEs that need task-specific automation but have limited resources for full custom robotics toolchains. A properly dimensioned 4-DOF palletizer can be significantly simpler and more affordable than a generic 6-DOF solution while still meeting operational requirements. Robodimm lowers the barrier to this adoption pathway by providing a practical sizing workflow that local integrators and in-house engineering teams can apply.

The case study also shows that mass-aware second-pass validation is not merely procedural: in the final scenario it changes the recommended J1 reduction while leaving the rest of the chain unchanged. This type of selective redesign signal is valuable in early-stage projects because it concentrates hardware changes where they matter, preserving architecture stability while improving robustness under payload and scale constraints.

\section{Code Availability}

Robodimm is available online at \url{https://customrobotics.es/}. The source code will be publicly available at \url{https://github.com/ual-arm/robodimm}. The DEMO version is open for public use. Access to PRO features can be arranged by contacting the developers.

\section{Conclusions}

This work presents Robodimm as a modular framework for simulation and dynamic analysis of industrial robotic manipulators, with emphasis on closed-chain mechanisms and actuator-aware design decisions. Its architecture allows users to move from interactive trajectory definition to physically grounded dynamic evaluation and actuator selection in a consistent pipeline.

\section*{Acknowledgements}

This work acknowledges the support and context provided by the LIFE ACCLIMATE initiative (LIFE20 CCA/ES/001824), \url{https://www.life-acclimate.eu/es/life-acclimate-2/}.

\bibliographystyle{unsrtnat}
\bibliography{references}

@article{BLANCOCLARACO2023101443,
title = {MultiVehicle Simulator (MVSim): Lightweight dynamics simulator for multiagents and mobile robotics research},
journal = {SoftwareX},
volume = {23},
pages = {101443},
year = {2023},
issn = {2352-7110},
doi = {https://doi.org/10.1016/j.softx.2023.101443},
author = {José-Luis Blanco-Claraco and Borys Tymchenko and Francisco José Mañas-Alvarez and Fernando Cañadas-Aránega and Ángel López-Gázquez and José Carlos Moreno},
}

@article{elsner2023panda,
  author = {Elsner, Jean},
  title = {Taming the Panda with Python: A powerful duo for seamless robotics programming and integration},
  journal = {SoftwareX},
  volume = {24},
  pages = {101532},
  year = {2023},
  doi = {10.1016/j.softx.2023.101532}
}

@article{barron2026mobile,
  author = {Barrón-Zambrano, José Hugo and Nuño-Maganda, Marco Aurelio and Hernández-Díaz, Melchor and Rangel-Magdaleno, José de Jesús and Hernández-Mier, Yahir},
  title = {Mobile2D-3D-RoboticSim: A robotic platform for computational thinking assessment in STEM and STEAM education},
  journal = {SoftwareX},
  volume = {33},
  pages = {102473},
  year = {2026},
  doi = {10.1016/j.softx.2025.102473}
}

@inproceedings{koenig2004gazebo,
  author = {Koenig, Nathan and Howard, Andrew},
  title = {Design and use paradigms for Gazebo, an open-source multi-robot simulator},
  booktitle = {Proceedings of the IEEE/RSJ International Conference on Intelligent Robots and Systems (IROS)},
  year = {2004},
  pages = {2149--2154},
  doi = {10.1109/IROS.2004.1389727}
}

@inproceedings{rohmer2013vrep,
  author = {Rohmer, Eric and Singh, Surya P. N. and Freese, Marc},
  title = {V-REP: A versatile and scalable robot simulation framework},
  booktitle = {Proceedings of the IEEE/RSJ International Conference on Intelligent Robots and Systems (IROS) Workshop},
  year = {2013}
}

@inproceedings{carpentier2018pinocchio,
  author = {Carpentier, Justin and Mansard, Nicolas},
  title = {Analytical derivatives of rigid body dynamics algorithms and their application to optimal control},
  booktitle = {Proceedings of Robotics: Science and Systems (RSS)},
  year = {2018}
}

@inproceedings{pinocchio2019,
  author = {Carpentier, Justin and Saurel, Guillaume and Buondonno, G{\'e}raldine and Mirabel, Joseph and Lamiraux, Florent and Stasse, Olivier and Mansard, Nicolas},
  title = {The Pinocchio C++ library: A fast and flexible implementation of rigid body dynamics algorithms and their analytical derivatives},
  booktitle = {Proceedings of the IEEE/SICE International Symposium on System Integration (SII)},
  year = {2019},
  pages = {614--619},
  doi = {10.1109/SII.2019.8700380}
}

@misc{pink_github,
  author = {Caron, St{\'e}phane and De Mont-Marin, Yann and Budhiraja, Rohan and Bang, Seung Hyeon and Domrachev, Ivan and Nedelchev, Simeon and Du, Peter and Escande, Adrien and Vaillant, Joris and Wingo, Bruce and Patapati, Santosh},
  title = {Pink: Python inverse kinematics based on Pinocchio},
  howpublished = {\url{https://github.com/stephane-caron/pink}},
  note = {Apache-2.0 license, version 4.0.0},
  year = {2026}
}

@article{macenski2022ros2,
  author = {Macenski, Steven and Foote, Tully and Gerkey, Brian and Lalancette, Christian and Woodall, William},
  title = {Robot Operating System 2: Design, architecture, and uses in the wild},
  journal = {Science Robotics},
  volume = {7},
  number = {66},
  year = {2022},
  doi = {10.1126/scirobotics.abm6074}
}

\end{document}